\documentclass[preprint,12pt]{elsarticle}

\usepackage[T1]{fontenc}
\usepackage[utf8]{inputenc}
\usepackage{lmodern}
\usepackage{amsmath,amssymb}
\usepackage{graphicx}
\usepackage{booktabs}
\usepackage{tabularx}
\usepackage{array}
\usepackage{subcaption}
\usepackage{xcolor}
\usepackage{enumitem}
\usepackage{siunitx}
\usepackage{hyperref}
\usepackage{url}

\journal{Forensic Science International: Digital Investigation}
\hypersetup{colorlinks=true,linkcolor=black,citecolor=black,urlcolor=blue}

\newcommand{\fake}{\mathrm{fake}}

\newcommand{\safeincludegraphics}[2][]{%
  \IfFileExists{#2}{\includegraphics[#1]{#2}}{%
  \fbox{\parbox[c][5.0cm][c]{0.90\linewidth}{\centering\textbf{Figure file to insert:}\\[0.3em]\texttt{#2}}}%
  }%
}

\begin{document}

\begin{frontmatter}

\title{\textsc{IRIS-GAN}: Staged Specialist Detection of Deepfake Faces}

\author[uv,ific]{Jaume Martinez Trenchs}
\author[uv,ific]{Ver\'onica Sanz\corref{cor1}}
\cortext[cor1]{Corresponding author}
\ead{veronica.sanz@ific.uv.es}

\address[uv]{Departamento de F\'isica Te\'orica, Universitat de Val\`encia, Burjassot, Spain}
\address[ific]{Instituto de F\'isica Corpuscular (IFIC), CSIC--Universitat de Val\`encia, Valencia, Spain}

\begin{abstract}
We introduce \textsc{IRIS-GAN}, a specialist forensic detector for
synthetic face images under cross-generator shift. Rather than addressing
universal synthetic-image detection, we focus on faces generated by
generative adversarial networks (GANs), which are state-of-the-art in deepfake content, and train the detector through staged
exposure to increasingly demanding GAN families while retaining earlier
generators.  The final model reaches
fake-detection rates above 99\% across the GAN families considered and
classifies an external real-face dataset with 98.9\% accuracy. Grad-CAM
analysis further reveals measurable generator-dependent spatial response
patterns, which remain informative for a secondary heatmap-only classifier.
Out-of-family tests on diffusion-generated faces confirm that
\textsc{IRIS-GAN} is a specialist detector, with some capability to reach non-GAN deepfakes.
These results establish staged training as an effective strategy for robust
GAN-face forensics.
\end{abstract}


\end{frontmatter}
\section{Introduction}
\label{sec:introduction}

The generation of photorealistic synthetic images has progressed rapidly in
recent years. Generative adversarial networks (GANs)~\cite{goodfellow2014gan}
established a powerful framework for image synthesis, and successive
architectures such as ProGAN~\cite{karras2018progan}, StyleGAN and its
variants~\cite{karras2019stylegan,karras2020stylegan2,karras2021stylegan3,
sauer2022styleganxl}, and EG3D~\cite{chan2022eg3d} have produced increasingly
realistic face images. In parallel, diffusion-based generators have expanded
the range and accessibility of high-quality synthetic content
~\cite{rombach2022latentdiffusion}. For digital forensics, this creates a
fundamental difficulty: the generative model encountered in practice may not
be represented among the models used to train a detector.

This limitation is now well documented. Detectors trained to discriminate
real images from fakes produced by a particular generator can reach very high
in-distribution performance while failing on images generated by unseen
architectures. Wang et al.~\cite{wang2020cnngenerated} showed that training
diversity and augmentation can substantially improve generalization from
CNN-generated images. More recently, Ojha et al.~\cite{ojha2023universal}
demonstrated that supervised real-versus-fake classifiers trained on one
generator family may map unseen synthetic images into the real class,
particularly when evaluated on newer generative paradigms. Large-scale
benchmarks such as GenImage~\cite{zhu2023genimage} further confirm that
cross-generator evaluation, rather than performance on a matched test set, is
central to assessing the practical reliability of AI-generated image
detectors.

\begin{figure}[h!]
\centering
\safeincludegraphics[width=\linewidth]{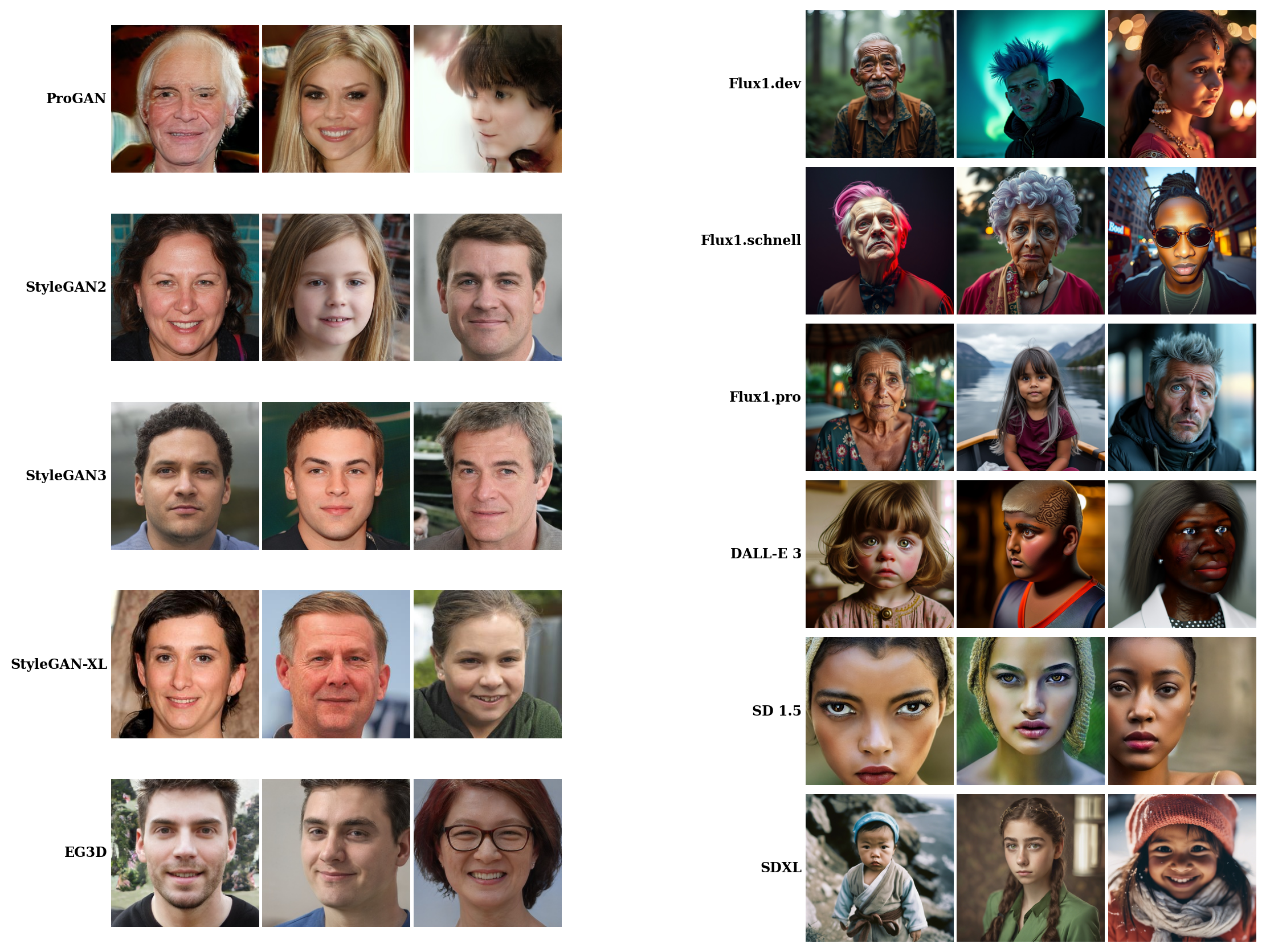}
\caption{Scope of the study. The main analysis concerns faces generated by several GAN families; diffusion-generated faces are used only as an out-of-family stress test in ~\ref{app:diffusion}.}
\label{fig:scope_examples}
\end{figure}

These observations motivate the deliberately restricted scope of this work.
We do not attempt to construct a universal detector for all synthetic images.
Instead, we study a coherent forensic setting: detection of
\emph{GAN-generated face images} across a sequence of related but increasingly
challenging generator families. This choice is not based on the assumption
that other forms of synthetic imagery are irrelevant or visually
distinguishable by inspection. Rather, it allows us to investigate a precise
question: whether the organization of training can improve transfer across
GAN generators that share a broad synthesis paradigm while differing
substantially in image quality, architecture, and learned artefacts.
Figure~\ref{fig:scope_examples} illustrates the real and synthetic face-image
domains considered in this study.
Within this setting, we focus on a practical failure mode observed during
model development. A conventional binary classifier trained simultaneously on
real images and several GAN families obtained excellent validation accuracy,
but generalized poorly to an unseen GAN family. We therefore propose a
\emph{staged specialist detector}: a detector trained through progressive
exposure to increasingly advanced GAN generators, while retaining previously
introduced fake families at later stages. The purpose of this curriculum is
not to increase in-distribution performance, which is already high, but to
reduce overspecialization to generator-specific cues and improve transfer to
GAN families not yet introduced at a given stage.

This specialist perspective also clarifies the role of negative and
out-of-family results. We initially explored reconstruction-based anomaly
detection, motivated by the possibility that a model trained on real faces
might identify synthetic images through increased reconstruction error.
However, in our GAN-face setting, reconstructed fake images were not
consistently separated from real ones. We therefore retain this study as a
negative methodological result in~\ref{app:reconstruction}, rather
than as part of the main detector. This conclusion is domain-specific:
reconstruction-based representations have been shown to be useful for
diffusion-generated image detection, for example through DIRE
~\cite{wang2023dire}. Similarly, we evaluate the final GAN specialist on
diffusion-generated faces only as an out-of-family stress test in~\ref{app:diffusion}. Previous studies have already shown that
detectors developed for GAN imagery may transfer poorly to diffusion outputs
~\cite{corvi2023diffusion,ojha2023universal}.

In addition to detection performance, we analyze the spatial response of the
final staged detector using Grad-CAM~\cite{selvaraju2017gradcam}. This
analysis has two purposes. First, it provides a diagnostic check against
trivial shortcut learning, such as persistent attention to a fixed facial
region or preprocessing artefact. Second, it allows us to test whether
heatmaps generated by the detector retain discriminative information across
GAN families. We therefore examine representative and averaged Grad-CAM
maps, quantify their spatial distributions, and train a secondary classifier
using heatmaps alone as input. We interpret these heatmaps conservatively:
saliency maps can reveal structured response patterns, but they do not by
themselves establish causal forensic evidence
~\cite{adebayo2018sanity}.

The main contribution of this work is to define and analyze a focused cross-generator forensic task for GAN-generated face images, using ProGAN, StyleGAN2, StyleGAN3, StyleGANXL, and EG3D as progressively challenging generator domains. Within this setting, we show that a one-shot real-versus-fake classifier can achieve strong matched-domain validation performance while generalizing poorly to an unseen GAN family. We then introduce a staged training protocol that retains earlier generators while incorporating more advanced ones, demonstrating in a controlled comparison that this strategy substantially improves transfer to unseen GANs. To interpret the resulting detector, we characterize its response through Grad-CAM averages, heatmap statistics, and effect-size comparisons, and further show that a secondary heatmap-only classifier preserves substantial GAN-detection ability. Finally, we report reconstruction-based and diffusion out-of-family tests as explicit limitations of the specialist approach, thereby defining its forensic domain of validity.

The remainder of this paper is organized as follows.
Section~\ref{sec:task_data} introduces the forensic task, datasets,
preprocessing, and evaluation criteria. Section~\ref{sec:methods} describes
the non-staged baseline, the staged specialist detector, and the heatmap-based
analysis. Section~\ref{sec:results} reports the main GAN-domain results and
the diagnostic heatmap study. Section~\ref{sec:discussion} discusses the
forensic interpretation and limitations of the approach.
\ref{app:reconstruction} reports the reconstruction-based experiments,
and \ref{app:diffusion} presents the out-of-family evaluation on
diffusion-generated images.

%

\section{Forensic task, datasets and evaluation protocol}
\label{sec:task_data}

\subsection{Task definition}

Given a face image $x$, the detector outputs a fake score $p_\theta(x)=P_\theta(y=\fake\mid x)$. The main task is binary classification between real images and GAN-generated face images. The main evaluation target is \emph{cross-generator generalization}: performance on GAN families not present at the relevant training stage.

Generated images are treated as the positive class. For a generator family $g$, we report the fake-detection rate
\begin{equation}
    \mathrm{FDR}_g = \frac{\mathrm{TP}_g}{\mathrm{TP}_g+\mathrm{FN}_g},
\end{equation}
which measures the fraction of fake images from family $g$ detected as fake. For real images, we report the false-positive rate
\begin{equation}
    \mathrm{FPR} = \frac{\mathrm{FP}}{\mathrm{FP}+\mathrm{TN}},
\end{equation}
where a false positive is a real image incorrectly labelled as fake. Confusion matrices are used to inspect error modes at the default threshold $\tau=0.5$. Threshold-independent ROC-AUC and precision--recall analyses are natural complementary metrics for a final forensic benchmark, but the central transfer results reported here use family-wise FDR and real-image FPR.

\subsection{Real and GAN-generated face datasets}

FFHQ~\cite{karras2019stylegan} is used as the real-image source for training the image-level detector. CelebA~\cite{liu2015celeba}, which differs in resolution and visual statistics, is used as an external real-image test. The GAN dataset comprises ProGAN, StyleGAN2, StyleGAN3, StyleGANXL and EG3D. All images are finally represented at $256\times256$ pixels. Table~\ref{tab:gan_sources} reports the GAN sources and their role in the experiments.

\begin{table}[t]
\centering
\caption{GAN-generated face datasets used in the study. ``Heatmap train'' denotes images used only for training the secondary heatmap classifier.}
\label{tab:gan_sources}
\small
\begin{tabularx}{\linewidth}{@{}llXXX@{}}
\toprule
Generator & Generation setting &  Image train/val & Test & Heatmap train \\
\midrule
ProGAN~\cite{karras2018progan} & $256^2$; random latent sampling & 14,000 & 1,000 & -- \\
StyleGAN2~\cite{karras2020stylegan2} & $1024^2$; truncation $\psi=0.7$ & 14,000 & 1,000 & 2,500 \\
StyleGAN3~\cite{karras2021stylegan3} & $1024^2$; truncation $\psi=0.7$ & 13,000 & 1,000 & 2,500 \\
StyleGANXL~\cite{sauer2022styleganxl} & $256^2$; truncation $\psi=0.7$ & 8,000 & 1,000 & 2,500 \\
EG3D~\cite{chan2022eg3d} & $512^2$; $\psi=0.7$, random viewpoint & 4,000 & 1,000 & 2,500 \\
\bottomrule
\end{tabularx}
\end{table}

For the image-level detector, 50,000 FFHQ images are used in the staged training pool with a 90/10 training/validation division at each stage. The external real-image evaluation uses 1,000 CelebA images. For the heatmap-only classifier, real heatmaps are generated from 10,000 CelebA images, which are not used to train the image-level detector. This separation is important for preventing leakage between the detector producing the heatmaps and the secondary classifier trained from them.

\subsection{Image preprocessing}

All images are processed at $256\times256$ pixels. During training, the same augmentation pipeline is applied at every stage: random resized crop covering 75--100\% of the image area with aspect ratio in $[0.9,1.1]$, horizontal flip with probability 0.5, JPEG re-encoding with probability 0.35 and quality sampled in $[55,95]$, mild colour jitter, grayscale conversion with probability 0.05, and Gaussian blur. At validation and test time, no stochastic augmentation is used: images are resized so that the shorter side is 256 pixels and centre-cropped. Inputs are normalized using ImageNet channel statistics.

\section{Methods}
\label{sec:methods}

\subsection{One-shot baseline and staged detector}
\label{subsec:staged}

The baseline is a conventional binary detector trained in a single step on real images together with the GAN images available at that point. It minimizes cross-entropy,
\begin{equation}
\mathcal{L}_{\mathrm{CE}}=-\frac{1}{N}\sum_{i=1}^{N}\left[y_i\log p_\theta(x_i)+(1-y_i)\log(1-p_\theta(x_i))\right],
\end{equation}
where $y_i=1$ indicates a generated image.

The staged detector uses the same binary objective, but the training distribution evolves. At stage $k$, newly introduced GAN images are added while earlier GAN families are retained:
\begin{equation}
    \mathcal{D}_k=\mathcal{R}_k\cup\bigcup_{j=1}^{k}\mathcal{G}_{j,k},
\end{equation}
where $\mathcal{R}_k$ denotes real training images and $\mathcal{G}_{j,k}$ the subset of GAN family $j$ retained at stage $k$. The protocol is summarized in Table~\ref{tab:stages}. Its purpose is to prevent forgetting of earlier generators while forcing the decision boundary to accommodate progressively less familiar GAN distributions.

\begin{table}[t]
\centering
\caption{Four-stage training protocol for the GAN-specialist detector. At every stage, earlier GAN families remain represented in the training set.}
\label{tab:stages}
\small
\begin{tabular}{@{}lrrrrrr@{}}
\toprule
Stage & FFHQ & ProGAN & StyleGAN2 & StyleGAN3 & StyleGANXL & EG3D \\
\midrule
S1 & 6,000  & 6,000 & 0     & 0     & 0     & 0 \\
S2 & 10,000 & 3,000 & 7,000 & 0     & 0     & 0 \\
S3 & 14,000 & 2,000 & 4,000 & 8,000 & 0     & 0 \\
S4 & 20,000 & 2,000 & 3,000 & 3,000 & 8,000 & 4,000 \\
\bottomrule
\end{tabular}
\end{table}

\subsection{Backbone and optimization}

The final detector is based on ConvNeXt-Large~\cite{liu2022convnext}, initialized with ImageNet-22k pretrained weights using the PyTorch Image Models implementation~\cite{wightman2019timm}. The original classification head is replaced by a two-class real/fake head. Early feature stages remain frozen; deeper features are progressively adapted as training advances. Table~\ref{tab:trainable_layers} summarizes the trainable components.

\begin{table}[t]
\centering
\caption{Trainable components of the ConvNeXt-Large detector at each stage.}
\label{tab:trainable_layers}
\small
\begin{tabularx}{\linewidth}{@{}lX@{}}
\toprule
Stage & Trainable components \\
\midrule
S1 & Classification head, fourth feature stage, final normalization layer \\
S2 & S1 components and last third of third feature stage \\
S3 & Classification head, full third and fourth feature stages, final normalization layer \\
S4 & Same trainable components as S3 \\
\bottomrule
\end{tabularx}
\end{table}

Stage S1 starts from the pretrained weights, while S2--S4 start from the best validation checkpoint of the preceding stage. All stages use AdamW with weight decay 0.05, learning rates $3\times10^{-4}$ for the classification head and $3\times10^{-5}$ for unfrozen backbone layers, batch size 64, and ten epochs. The best validation checkpoint is retained. Training is performed on a single NVIDIA Tesla V100-PCIe-32GB GPU; the complete four-stage procedure takes approximately 1 h 42 min.

\subsection{Representation and heatmap diagnostics}
\label{subsec:heatmaps}

We use two diagnostics for the final staged detector. First, we compare the internal representations associated with different domains. For each image, we extract the feature vector before the final classification head and compute a mean vector $\mu_d$ for each domain $d$. The domain similarity is the Pearson correlation
\begin{equation}
    S_{ab}=\mathrm{corr}_{\mathrm{Pearson}}(\mu_a,\mu_b),
\end{equation}
shown as a correlation matrix.

Second, we compute Grad-CAM maps~\cite{selvaraju2017gradcam} using the last ConvNeXt feature stage. Individual maps are min--max normalized to $[0,1]$ after bilinear resizing to the input resolution. For each domain, average heatmaps are computed from 1,000 normalized maps. We characterize the maps through entropy, centre-of-mass radius, central energy, left--right and top--bottom asymmetry, and high-activation area. Differences relative to real images are summarized with Cliff's delta~\cite{cliff1993dominance}.

To assess whether these heatmaps contain discriminative information beyond visual inspection, we train a ResNet18~\cite{he2016resnet} binary classifier using only Grad-CAM maps. Its training set contains 10,000 CelebA-derived real heatmaps and 10,000 heatmaps from StyleGAN2, StyleGAN3, StyleGANXL and EG3D, balanced across GAN families. ProGAN is excluded from heatmap training and retained as a transfer test.

\section{Results}
\label{sec:results}

\subsection{Staged training improves unseen-GAN transfer}

We first isolate the effect of the training strategy. Both one-shot and staged detectors are trained using real images and GAN families up to StyleGAN2, and then tested on StyleGAN3, which is not included in that comparison's training set. As shown in Table~\ref{tab:staged_vs_nonstaged}, the one-shot model reaches near-saturated validation accuracy but detects only approximately 54\% of unseen StyleGAN3 images. Staged training lowers accuracy on the familiar validation distribution but increases StyleGAN3 detection to above 80\%. This is the central evidence that validation accuracy on known generators is not a sufficient forensic criterion.

\begin{table}[t]
\centering
\caption{Controlled comparison of one-shot and staged training. StyleGAN3 is unseen during this experiment. Values are fake-detection accuracies at the default threshold.}
\label{tab:staged_vs_nonstaged}
\small
\begin{tabular}{@{}lccc@{}}
\toprule
Model & Validation accuracy & StyleGAN2 & Unseen StyleGAN3 \\
\midrule
One-shot, 10 epochs & 0.9900 & 0.999 & 0.540 \\
One-shot, 20 epochs & 0.9922 & 0.998 & 0.547 \\
Staged, 10 epochs & 0.8916 & 0.901 & 0.835 \\
Staged, 20 epochs & 0.9057 & 0.901 & 0.812 \\
\bottomrule
\end{tabular}
\end{table}

\begin{figure}[h!]
\centering
\safeincludegraphics[width=\linewidth]{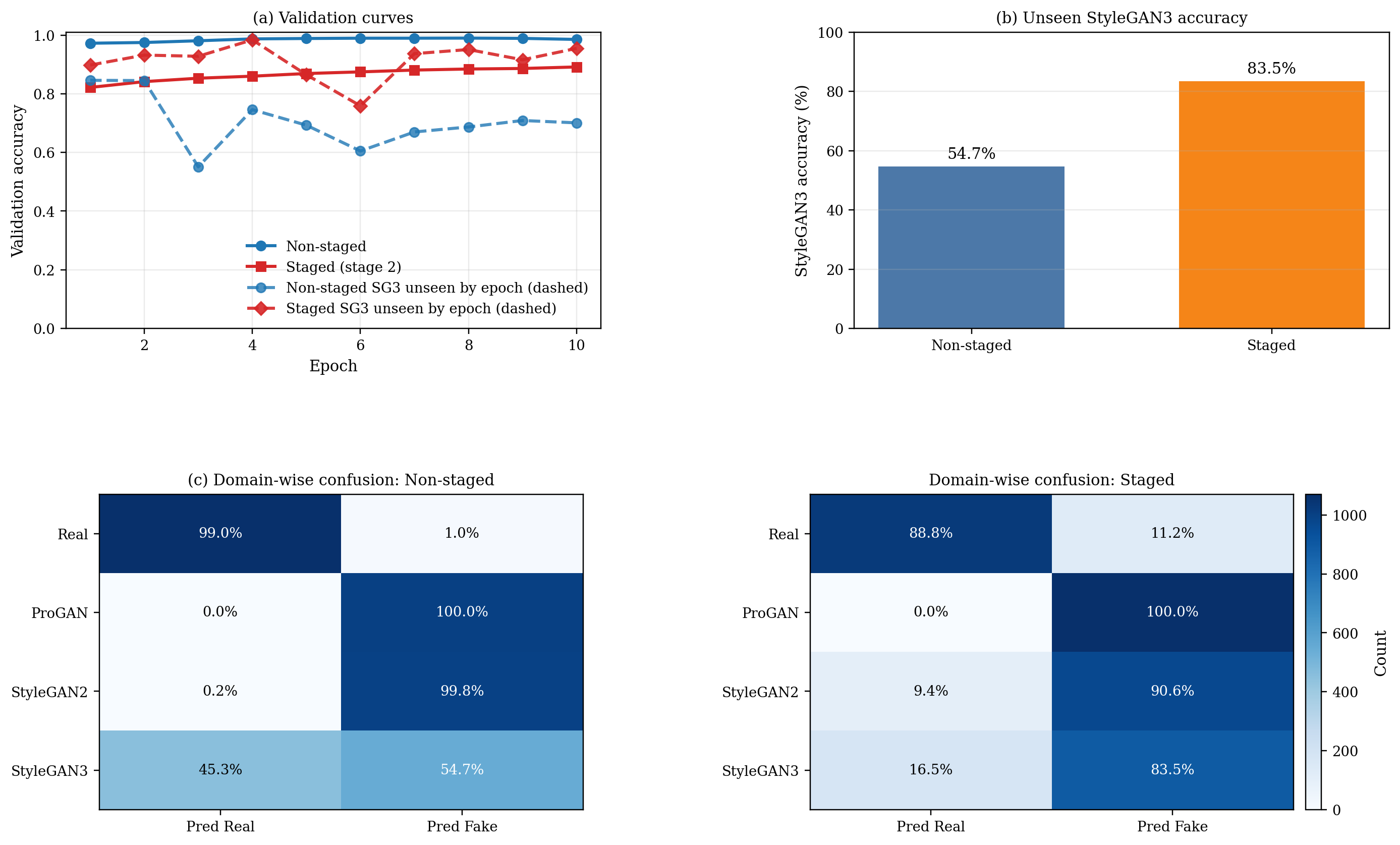}
\caption{Comparison between one-shot and staged training in the preliminary ProGAN/StyleGAN2 experiment. Despite lower validation accuracy on seen distributions, staged training transfers substantially better to unseen StyleGAN3 images.}
\label{fig:staged_vs_nonstaged}
\end{figure}

\subsection{The final staged detector covers the considered GAN families}

\begin{figure}[h!]
\centering
\safeincludegraphics[width=\linewidth]{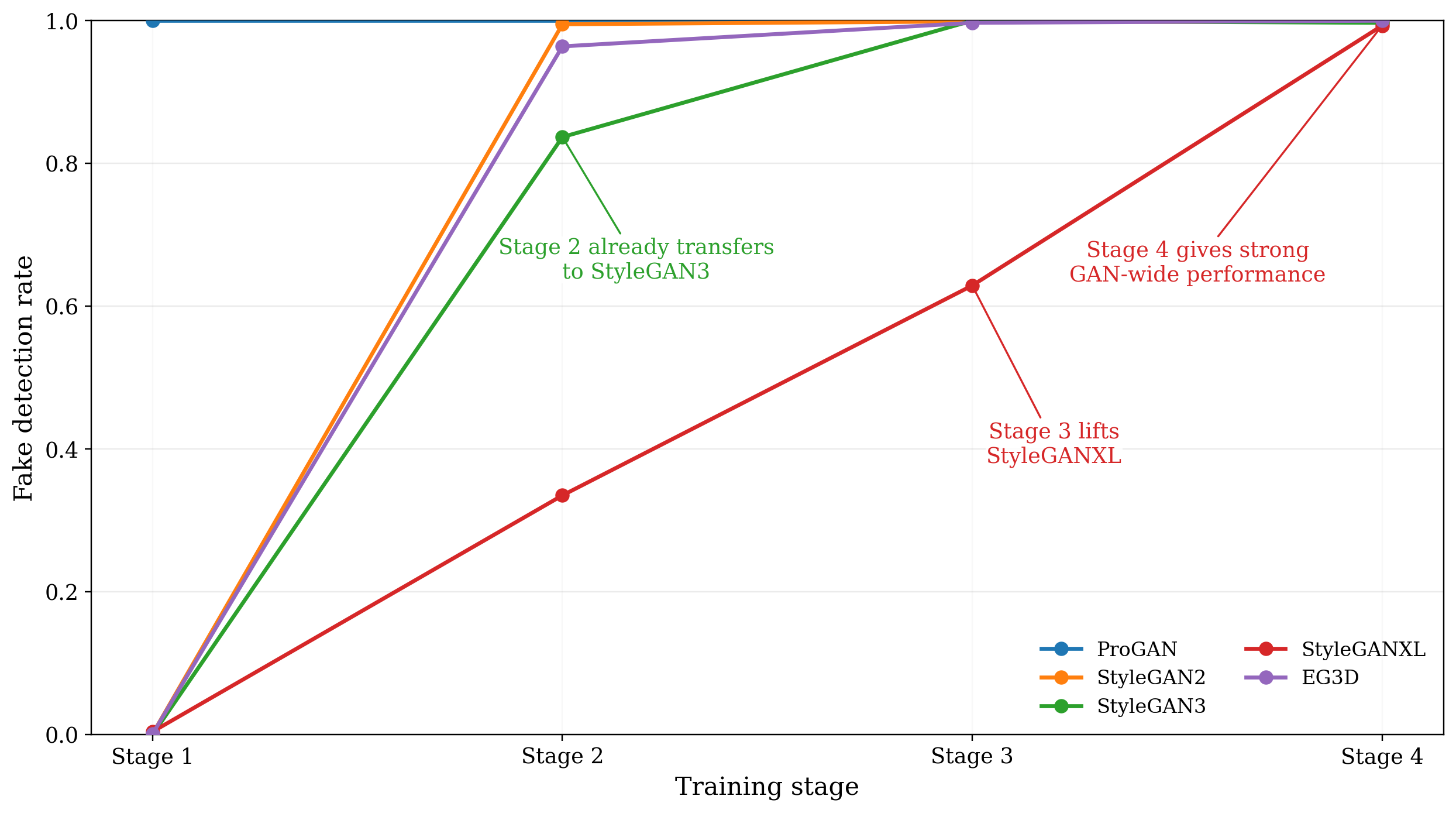}
\caption{Fake-detection rate across the four stages for each GAN family. StyleGAN3 already transfers strongly at S2, while StyleGANXL requires later explicit exposure.}
\label{fig:stage_evolution}
\end{figure}

Table~\ref{tab:stage_evolution} and Figure~\ref{fig:stage_evolution} show the evolution of family-wise fake-detection rates. S1 recognizes ProGAN but transfers poorly. After StyleGAN2 is introduced at S2, the detector already identifies 83.7\% of unseen StyleGAN3 images and transfers almost completely to EG3D. StyleGANXL remains the most difficult out-of-stage generator and benefits from explicit inclusion at S4. The final staged detector reaches fake-detection rates of at least 99.3\% on all GAN test families. On the external real-image set, 98.9\% of CelebA images are correctly classified as real, corresponding to a 1.1\% false-positive rate.

\begin{table}[t]
\centering
\caption{Fake-detection rate on held-out GAN test sets after each stage. A low value means that images from that generator are generally misclassified as real.}
\label{tab:stage_evolution}
\small
\begin{tabular}{@{}lccccc@{}}
\toprule
Stage & ProGAN & StyleGAN2 & StyleGAN3 & StyleGANXL & EG3D \\
\midrule
S1 & 1.000 & 0.001 & 0.000 & 0.004 & 0.001 \\
S2 & 1.000 & 0.995 & 0.837 & 0.335 & 0.964 \\
S3 & 1.000 & 0.999 & 1.000 & 0.629 & 0.997 \\
S4 & 1.000 & 0.998 & 0.997 & 0.993 & 1.000 \\
\bottomrule
\end{tabular}
\end{table}

\subsection{Representation similarity and Grad-CAM diagnostics}

\begin{figure}[h!]
\centering
\safeincludegraphics[width=0.82\linewidth]{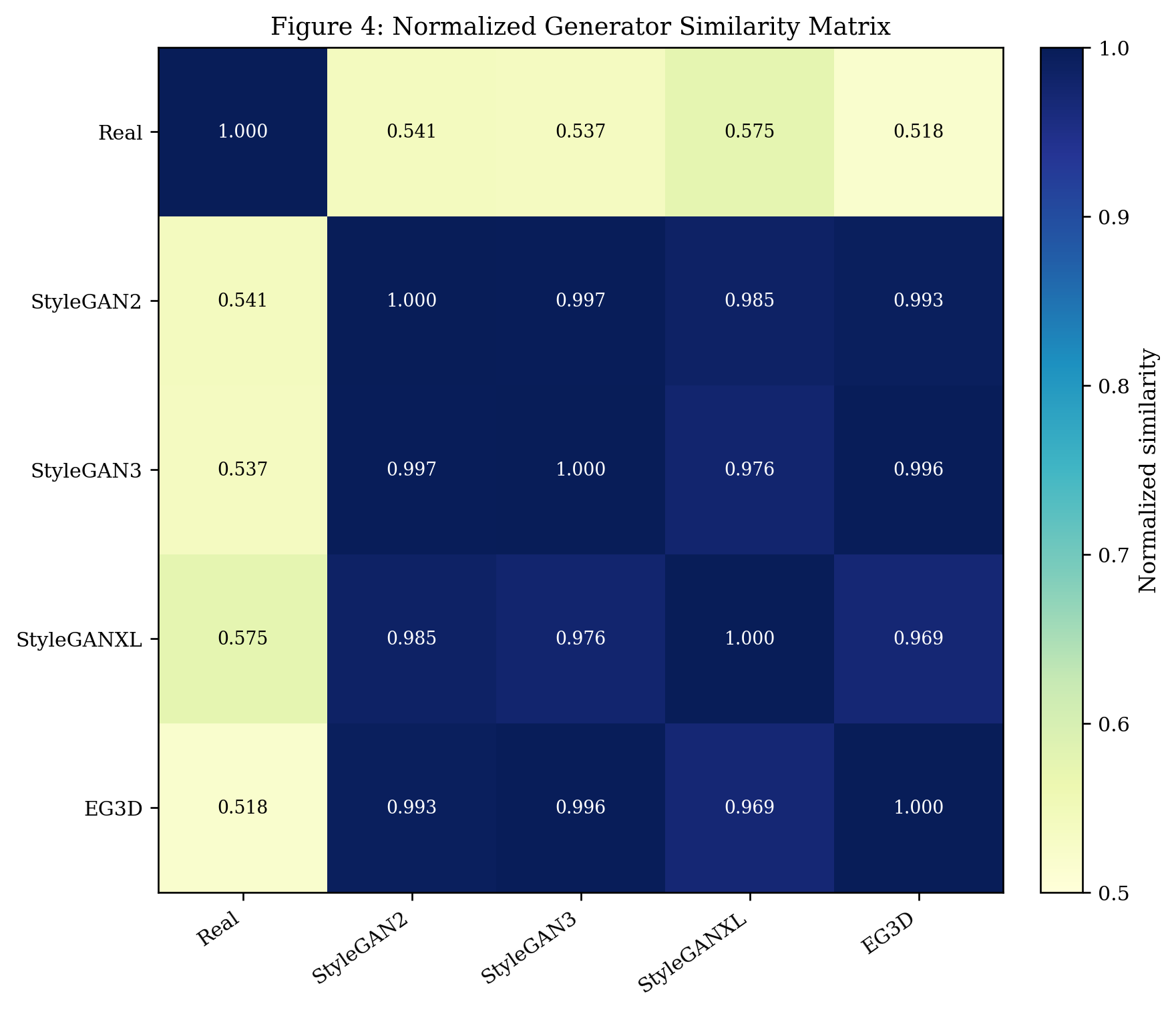}
\caption{Correlation matrix between domain-mean ConvNeXt representations for the final staged detector. High correlations among StyleGAN2, StyleGAN3 and EG3D are consistent with the observed cross-generator transfer pattern.}
\label{fig:embedding_matrix}
\end{figure}

The embedding matrix in Figure~\ref{fig:embedding_matrix} provides a diagnostic view of the final detector. Within the plotted domains, StyleGAN2, StyleGAN3 and EG3D have highly correlated mean representations, while StyleGANXL remains moderately separated. This observation is consistent with the transfer pattern in Table~\ref{tab:stage_evolution}: StyleGAN3 generalizes early, while StyleGANXL requires explicit training exposure. Because similarity is computed only from domain means, this matrix is an interpretation tool rather than an independent test of performance.

\begin{figure}[h!]
\centering
\safeincludegraphics[width=\linewidth]{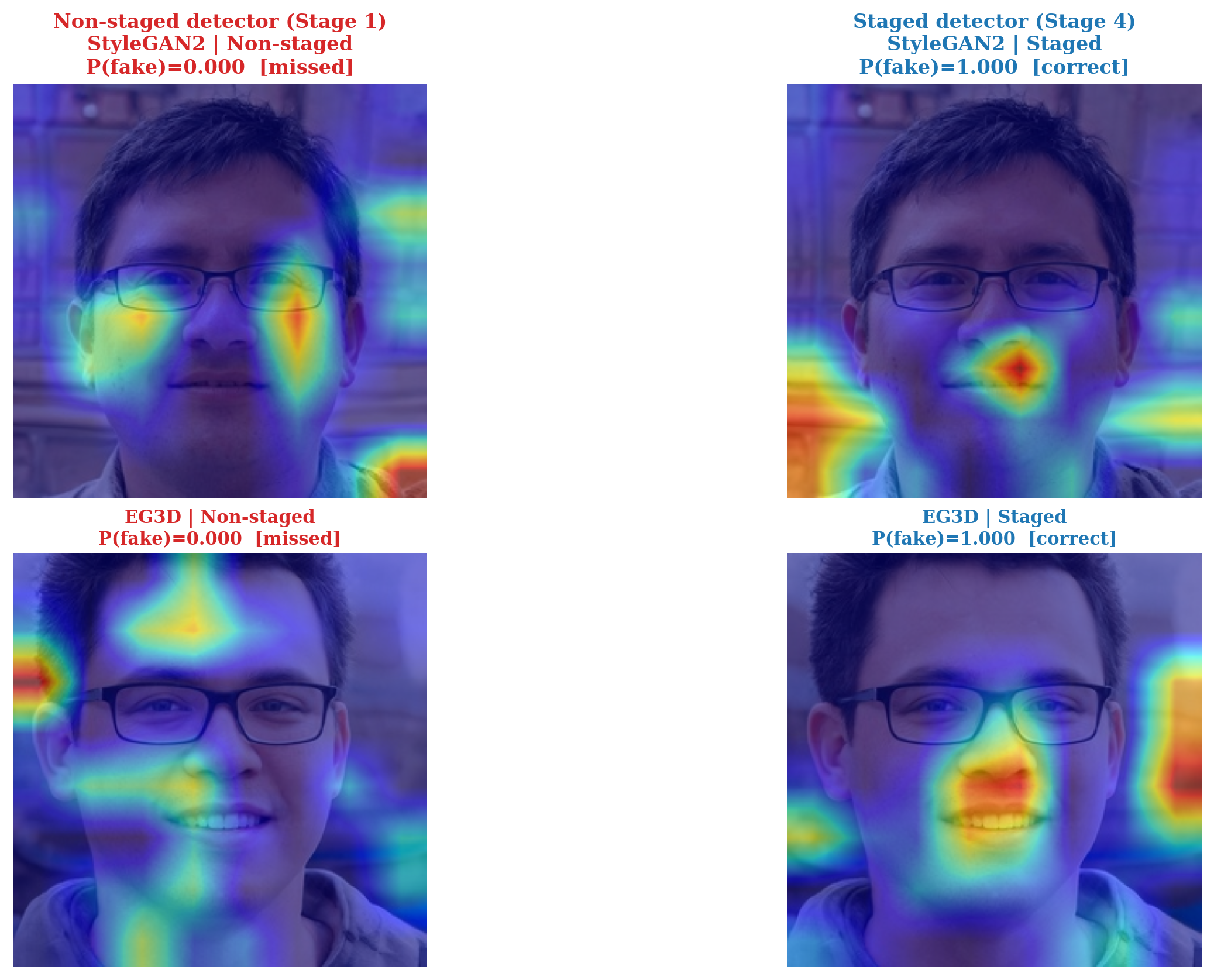}
\caption{Representative Grad-CAM maps comparing an early non-staged detector with the final staged detector. The maps visualize spatial response patterns associated with individual predictions.}
\label{fig:gradcam_examples}
\end{figure}

Grad-CAM maps provide a complementary spatial diagnostic. Figure~\ref{fig:gradcam_examples} contrasts representative responses from an earlier non-staged detector and the final staged detector. The examples suggest that the later detector is less dependent on a single localized response pattern; however, the maps are interpreted only as model-response visualizations, not as proof of causal forensic evidence.

\begin{figure}[h!]
\centering
\safeincludegraphics[width=\linewidth]{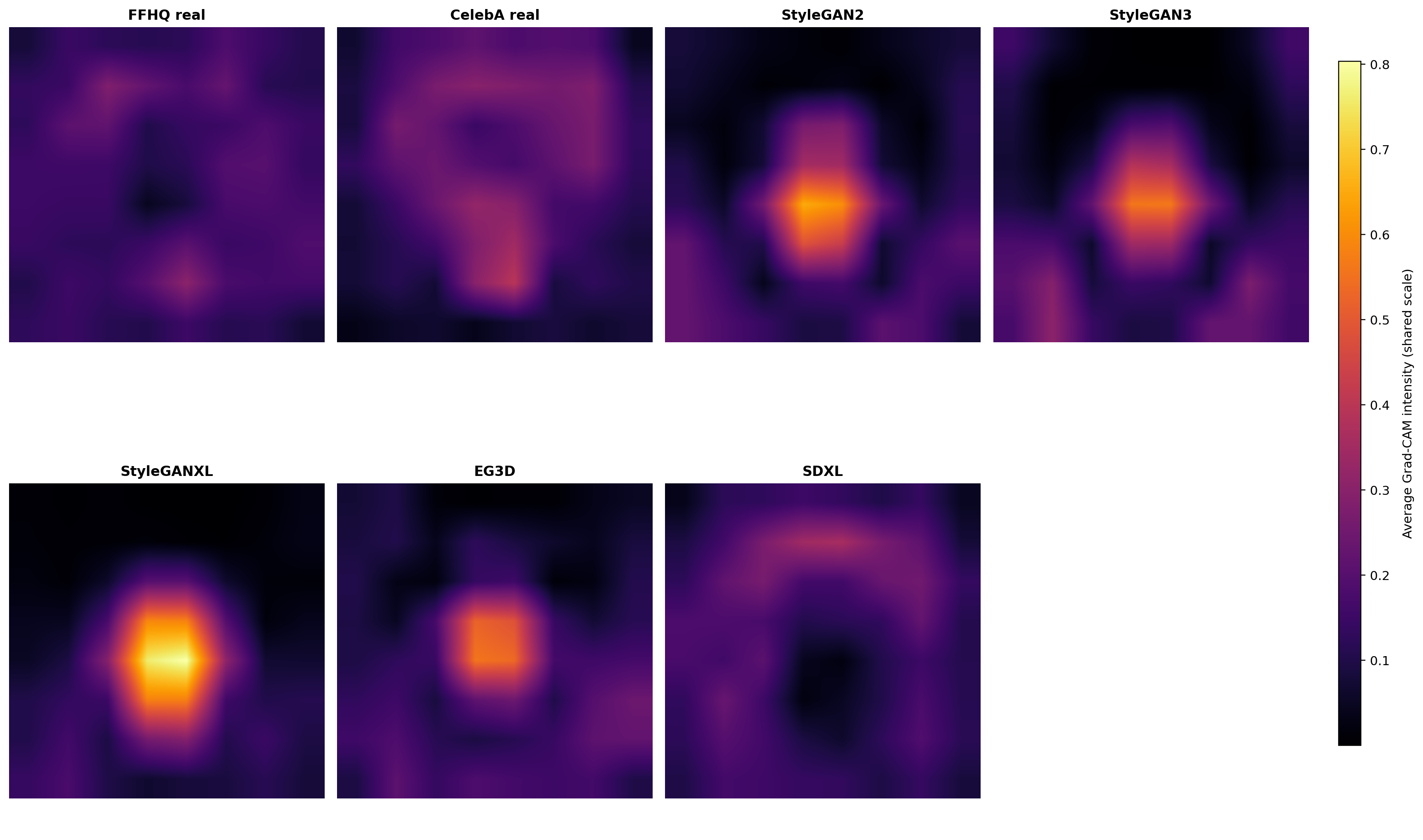}
\caption{Average Grad-CAM maps computed from 1,000 images per domain. The maps reveal domain-dependent spatial response patterns; SDXL is shown only as an out-of-family visual reference.}
\label{fig:average_heatmaps}
\end{figure}

\begin{figure}[h!]
\centering
\safeincludegraphics[width=\linewidth]{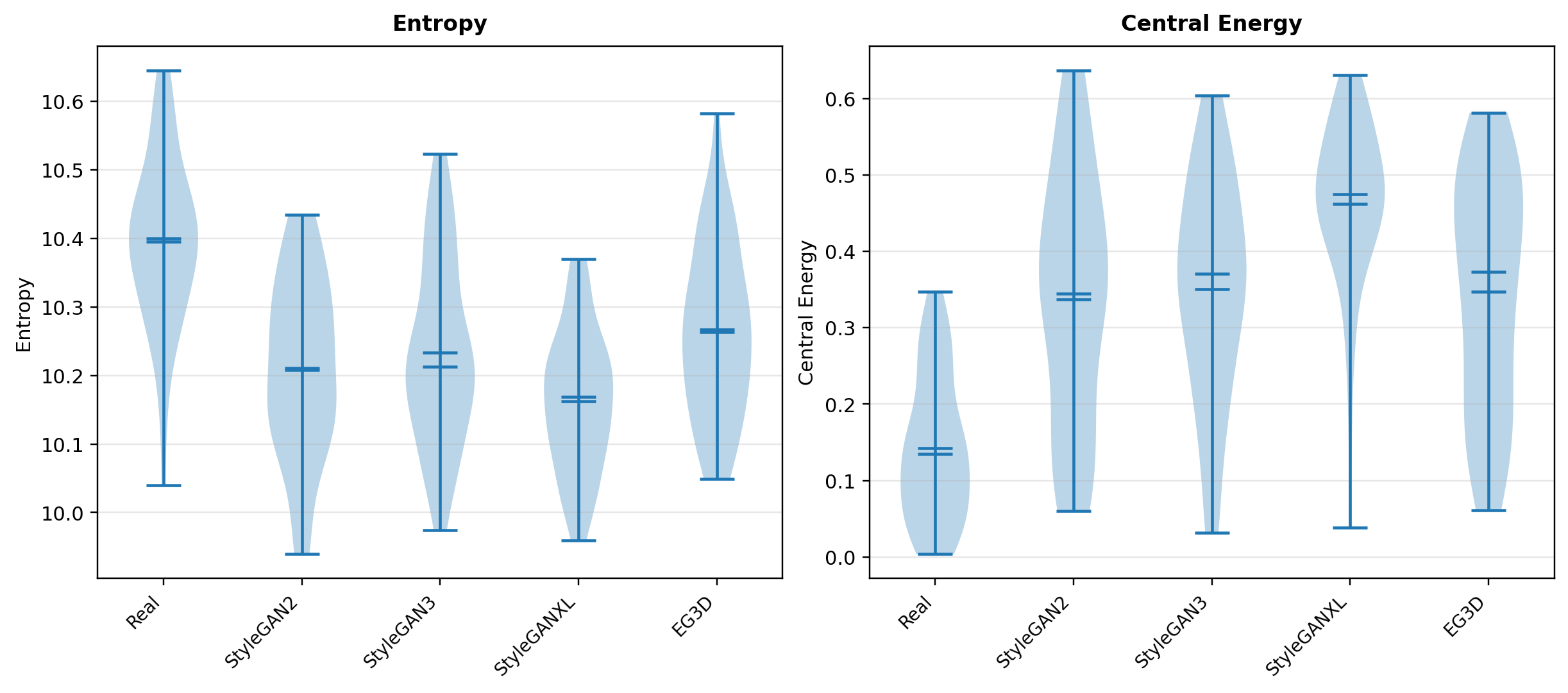}
\caption{Distributions of selected Grad-CAM statistics across real and GAN-generated domains. The shown variables illustrate differences in saliency concentration rather than causal image artifacts.}
\label{fig:heatmap_statistics}
\end{figure}

\begin{figure}[h!]
\centering
\safeincludegraphics[width=\linewidth]{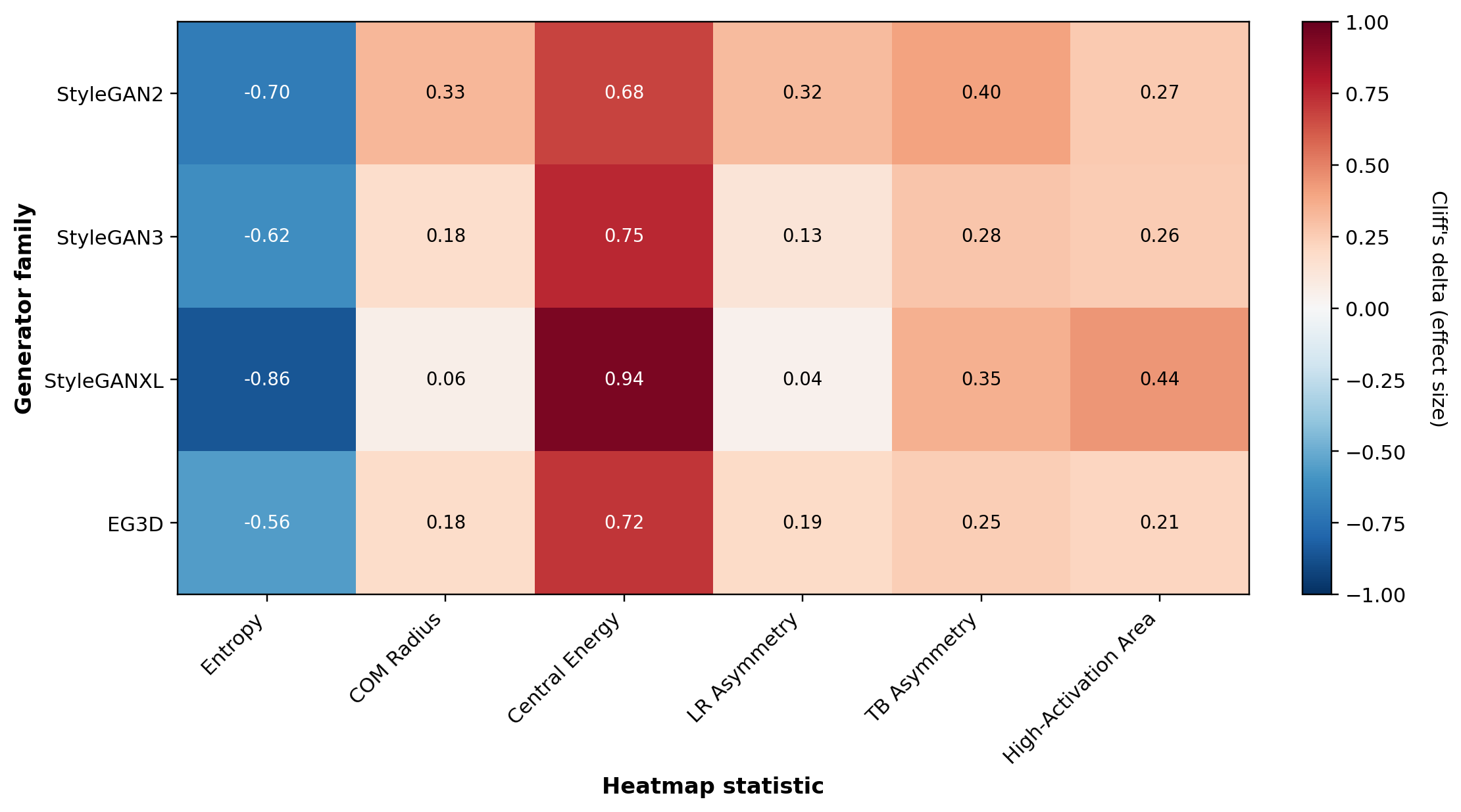}
\caption{Cliff's delta of heatmap statistics relative to the real-image baseline. The structured effect sizes indicate that different GAN families shift the detector response in distinct ways.}
\label{fig:cliffs_delta}
\end{figure}

Average maps in Figure~\ref{fig:average_heatmaps} and quantitative summaries in Figures~\ref{fig:heatmap_statistics} and~\ref{fig:cliffs_delta} show that saliency distributions vary systematically between real and GAN-generated domains. In particular, the Cliff's delta map identifies generator-dependent shifts in entropy, central energy and asymmetry-related variables. These patterns justify testing heatmaps as a compressed diagnostic representation.

\subsection{Heatmap-only classification retains substantial GAN information}

 The image-level detector used to produce the Grad-CAM maps was not trained on the images entering the heatmap-classifier training set. Real heatmaps are obtained from CelebA, while the fake heatmaps used for training are obtained from separate samples of StyleGAN2, StyleGAN3, StyleGANXL and EG3D. ProGAN is excluded from heatmap-classifier training and retained as an external GAN-family transfer test. Table~\ref{tab:heatmap_protocol} summarizes this protocol.

\begin{table}[t]
\centering
\caption{Protocol for the heatmap-only classifier. The image-level detector used to generate Grad-CAM maps was not trained on the images entering the heatmap-classifier training set.}
\label{tab:heatmap_protocol}
\small
\begin{tabular}{@{}llr@{}}
\toprule
Role & Dataset / source & Images \\
\midrule
\multicolumn{3}{@{}l}{\textit{Heatmap-classifier training set}} \\
Real heatmaps & CelebA & 10,000 \\
Fake heatmaps & StyleGAN2 & 2,500 \\
              & StyleGAN3 & 2,500 \\
              & StyleGANXL & 2,500 \\
              & EG3D & 2,500 \\
              & ProGAN & Excluded \\
\midrule
\multicolumn{3}{@{}l}{\textit{Evaluation set}} \\
Real images & FFHQ, disjoint split & 1,000 \\
Real images & CelebA, disjoint split & 1,000 \\
Fake images & Five GAN domains & 1,000 each \\
\bottomrule
\end{tabular}
\end{table}

Table~\ref{tab:heatmap_classifier} reports the  heatmap-classifier test. The classifier performs strongly on StyleGAN2, StyleGAN3, StyleGANXL and EG3D, despite receiving only Grad-CAM maps rather than original images. Its poor transfer to ProGAN, which was excluded from heatmap training, shows that heatmap signatures are not by themselves universal across GAN families. The result therefore complements rather than replaces the staged image detector.

\begin{table}[t]
\centering
\caption{Fake-detection rate of the heatmap-only classifier. ProGAN is not included in heatmap training.}
\label{tab:heatmap_classifier}
\small
\begin{tabular}{@{}lc@{}}
\toprule
Generator family & Heatmap-only fake-detection rate \\
\midrule
StyleGAN2 & 0.982 \\
StyleGAN3 & 0.966 \\
StyleGANXL & 0.964 \\
EG3D & 0.913 \\
ProGAN (unseen in heatmap training) & 0.480 \\
\bottomrule
\end{tabular}
\end{table}

\section{Discussion and conclusion}
\label{sec:discussion}

This study shows that a staged detector specialized in GAN-generated face images can generalize across the GAN families more reliably than an otherwise similar one-shot training strategy.The central comparison is not the final near-saturated result obtained after all GAN families have been introduced, but the intermediate transfer result: staged exposure to early GAN models substantially improves detection of unseen next-generation GAN images, suggesting that learning from multiple GAN styles can help the detector acquire features that remain informative across generator families and may therefore provide some preparedness for future GAN improvements.

The heatmap analysis strengthens the forensic reading of the detector while also imposing caution. Average Grad-CAM maps, summary statistics and the heatmap-only classifier all indicate that model response patterns contain domain-dependent information. These findings can be useful for diagnostic reporting and for identifying failure modes. They do not establish that a highlighted image region is a causal or human-interpretable trace of manipulation.

The detector is not a universal synthetic-image classifier. ~\ref{app:diffusion} shows that the GAN-specialist model does not transfer effectively to diffusion-generated faces, but also that the heatmap analysis is more sensitive to out of distribution than the image-based analysis. This limitation is operationally important: a practical synthetic-media system would require broader training data or a modular architecture with specialist components and explicit domain-of-validity reporting. Similarly, a detector intended for forensic deployment should be assessed under image compression, resizing, screenshots and realistic online post-processing, and should report calibration and false-positive rates under different operating thresholds.

Within its stated scope, however, the study demonstrates that the organization of training data matters. Retaining earlier GAN families while progressively introducing harder generators provides a simple and effective strategy for cross-generator robustness in GAN-face forensics. Heatmap-based diagnostics add an interpretable, though non-causal, view of the learned response and open a route toward more transparent specialist detectors.

\appendix
\section{Reconstruction-based anomaly detection}
\label{app:reconstruction}

Before adopting supervised GAN detection, we explored reconstruction-based anomaly detection, motivated by the idea that a model trained on real images might reconstruct synthetic faces poorly. This direction is conceptually related to anomaly-aware representation learning~\cite{khosa2023anomalyawareness}, although the experiments reported here focus on reconstruction error rather than the full anomaly-awareness framework.

We evaluated a frozen pretrained Stable Diffusion VAE, a partially fine-tuned version of the same VAE on real FFHQ faces, and a custom real-only VAE. Reconstruction differences were assessed through pixel-level L1 errors and perceptual VGG-feature distances; the custom VAE was also tested with and without KL regularization. In all cases, real and generated distributions overlap strongly. GAN-generated faces are not consistently reconstructed with larger error than real faces, and can even yield lower reconstruction error, plausibly because they are more standardized in pose, lighting or texture.

\begin{figure}[h!]
\centering
\safeincludegraphics[width=\linewidth]{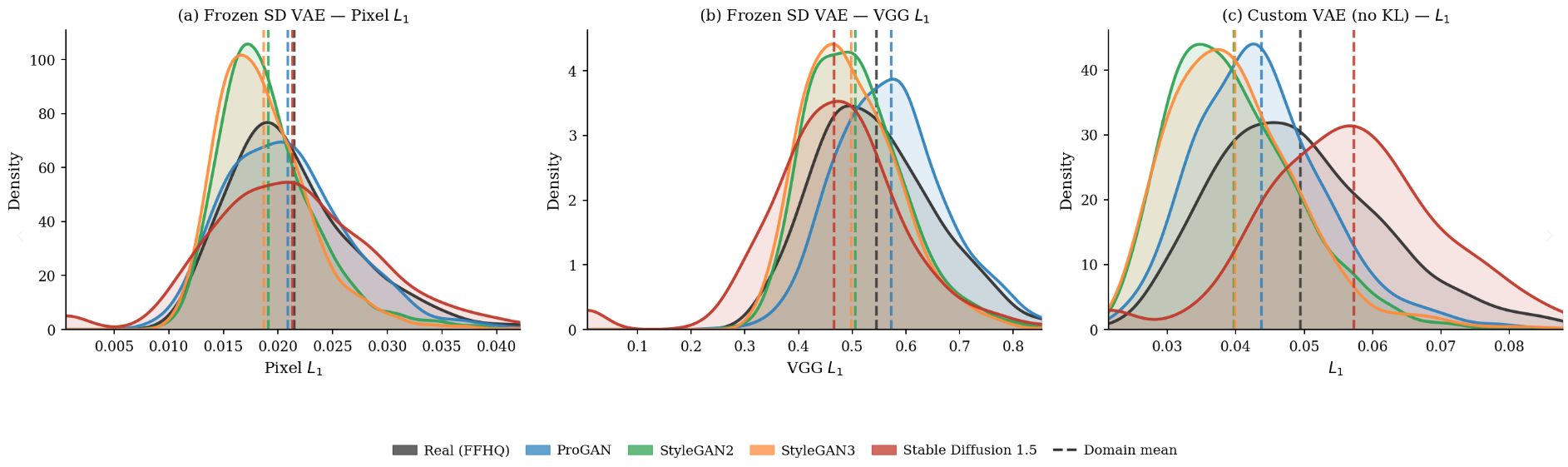}
\caption{Reconstruction-error distributions for 1,000 images per domain. Pixel and VGG-feature discrepancies from VAE reconstructions exhibit substantial real/fake overlap; a custom real-only VAE likewise fails to yield a clean separation signal.}
\label{fig:reconstruction}
\end{figure}

Figure~\ref{fig:reconstruction} summarizes the key negative result. Reconstruction error does not provide a sufficiently clean forensic discriminator for the present problem, and this motivated the staged supervised strategy used in the main paper.

\section{Out-of-family evaluation on diffusion-generated faces}
\label{app:diffusion}

Diffusion-generated faces are used as an out-of-family stress test, not as part of the main GAN-detection task. Stable Diffusion 1.5 images were generated locally, while FLUX.1-dev, FLUX.1-schnell, FLUX.1-pro, Stable Diffusion XL and DALL-E 3 images were selected from the SFHQ-T2I collection. One thousand images from each domain were evaluated.

\begin{figure}[h!]
\centering
\safeincludegraphics[width=\linewidth]{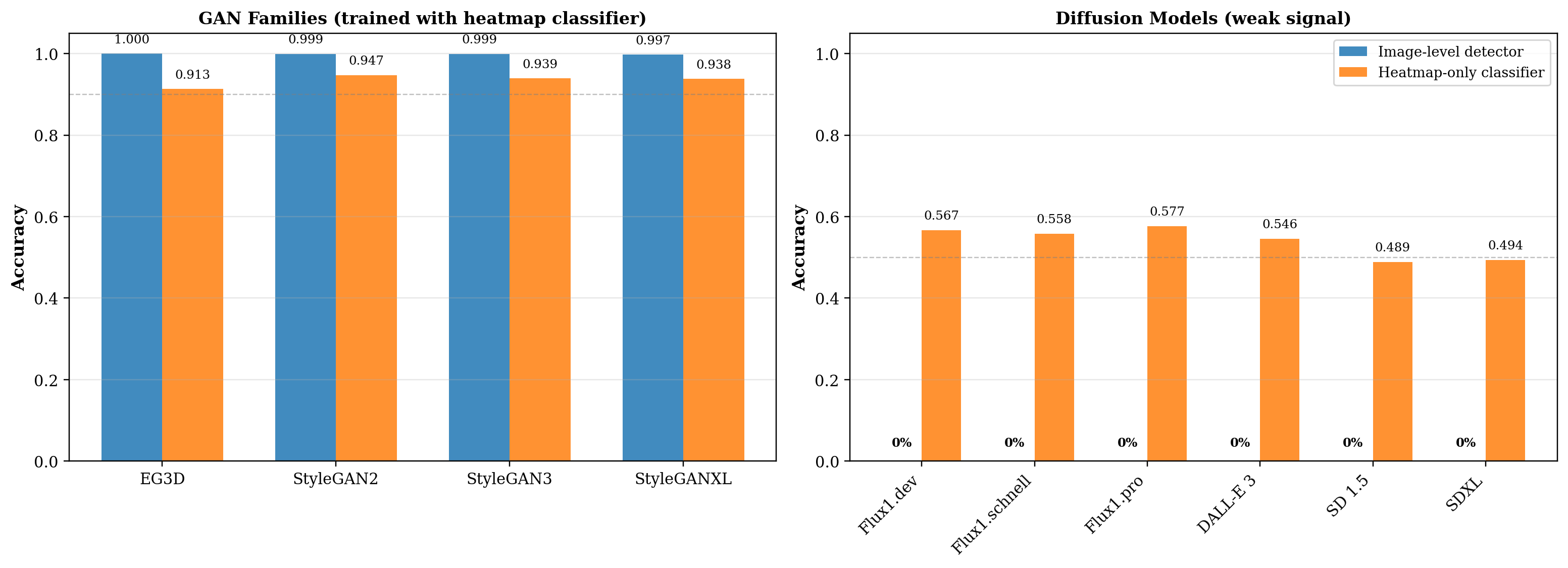}
\caption{Image-level staged detector and heatmap-only classifier evaluated on GAN and diffusion domains. The heatmap representation preserves useful GAN information but provides only weak out-of-family sensitivity to diffusion-generated faces.}
\label{fig:image_vs_heatmap}
\end{figure}

The staged image detector classifies essentially all diffusion-generated images as real, confirming that GAN-specialist training does not transfer automatically to a different generative mechanism. The heatmap-only classifier produces a weak nonzero signal for several diffusion models, but its detection rates remain unsuitable for forensic use. Lowering the decision threshold increases fake recall at the cost of incorrectly labelling a substantial fraction of real images as fake. Table~\ref{tab:diffusion_heatmaps} and Figure~\ref{fig:image_vs_heatmap} report this negative stress test.

\begin{table}[t]
\centering
\caption{Fake-detection rate of the heatmap-only classifier on diffusion-generated faces. The image-level GAN-specialist detector gives approximately zero detection rate on these domains.}
\label{tab:diffusion_heatmaps}
\small
\begin{tabular}{@{}lcc@{}}
\toprule
Diffusion domain & Default threshold & $\tau=0.3147$ \\
\midrule
FLUX.1-dev & 0.221 & 0.335 \\
FLUX.1-schnell & 0.205 & 0.305 \\
FLUX.1-pro & 0.241 & 0.335 \\
Stable Diffusion 1.5 & 0.067 & -- \\
Stable Diffusion XL & 0.076 & -- \\
DALL-E 3 & 0.181 & 0.284 \\
\bottomrule
\end{tabular}
\end{table}

\section*{Data and code availability}

The study uses public real-face datasets and images generated from publicly documented generative models. Code, trained weights and derived analysis products can be made available subject to dataset and model licensing conditions. For enquiries, contact veronica.sanz@uv.es.

\section*{Declaration of competing interest}

The authors declare that they have no known competing financial interests or personal relationships that could have appeared to influence the work reported in this paper.

\section*{CRediT authorship contribution statement}

\textbf{Jaume Martinez Trenchs:} Software, investigation, data curation, visualization, formal analysis, writing. 
\textbf{Ver\'onica Sanz:} Conceptualization, methodology, supervision, writing.


\end{document}